\documentclass[journal]{IEEEtran}
\usepackage{cite}
\usepackage{amsmath,amssymb,amsfonts}
\usepackage{verbatim}
\usepackage{graphicx}
\usepackage{textcomp}
\usepackage{xcolor}
\usepackage{times}
\usepackage{soul}
\usepackage{url}
\usepackage{graphicx}
\usepackage{amsmath}
\usepackage{subfigure}
\usepackage{float}
\usepackage{bm}
\usepackage{hyperref}
\usepackage{amsmath,amssymb,amsfonts,amsthm}
\usepackage{booktabs}
\usepackage{graphicx}
\usepackage{amssymb}
\usepackage{amsmath}
\usepackage{amsthm}
\usepackage{booktabs}
\usepackage[ruled]{algorithm2e} 
\usepackage{algorithmicx}
\usepackage{algpseudocode} 
\usepackage[misc]{ifsym}
\usepackage{graphicx}
\usepackage{subfigure}
\usepackage{color,xcolor}
\ifCLASSINFOpdf
\else
\fi
\begin{document}

%
\title{
TODE-Trans: \textbf{T}ransparent \textbf{O}bject \textbf{D}epth \\ \textbf{E}stimation with Transformer}
%

%
%

\author{ Kang Chen*\textsuperscript{1}, Shaochen Wang*\textsuperscript{1}, Beihao Xia\textsuperscript{2}, Dongxu Li\textsuperscript{1}, Zhen Kan\textsuperscript{1}, and Bin Li\textsuperscript{1} 
\thanks{* Contribute equally}
\thanks{\textsuperscript{1} University of
	Science and Technology of China, Hefei, China}
\thanks{\textsuperscript{2} Huazhong University of Science and Technology, Wuhan, China}
\thanks{Corresponding author: Zhen Kan (zkan@ustc.edu.cn)}
}

%
%

\maketitle
\thispagestyle{empty}
\pagestyle{empty}

%



\begin{abstract}
	
Transparent objects   are widely used in industrial automation and daily life.  However, robust visual recognition and perception of transparent objects have always been a major challenge.	
Currently, most commercial-grade depth cameras  are still not good at sensing the surfaces of transparent objects due to the refraction and reflection of light. 
In this work, we present a  transformer-based  transparent object  depth estimation approach   from a single RGB-D input. We observe that the global characteristics of the transformer make it easier to extract contextual information to perform depth estimation of transparent areas.
In addition, to better enhance the fine-grained features,  a feature fusion module (FFM) is designed to assist coherent prediction. 
Our empirical evidence demonstrates that our model delivers significant improvements  in recent popular datasets, e.g., $25\%$ gain on RMSE and $21\%$ gain on REL compared to previous state-of-the-art convolutional-based counterparts in ClearGrasp dataset.
Extensive results show that our transformer-based model enables better aggregation of the object's RGB and inaccurate depth information to  obtain a better depth representation.
Our code and the pre-trained model will be  available at 
{\url{https://github.com/yuchendoudou/TODE }.}
\end{abstract}


%
\IEEEpeerreviewmaketitle

\section{INTRODUCTION}
A large volume of transparent objects such as glasses, plastic bottles are frequently appearing in the household, manufacturing, and daily life. The study of computer vision on transparent objects has also attracted  a great deal of attention, including transparent area segmentation \cite{DBLP:conf/cvpr/KalraTRVRK20},\cite{guo2019transparent},  3D reconstruction\cite{DBLP:conf/cvpr/HanWL15}, especially for domestic service robots or industrial transparent object manipulation \cite{9197518}.
Unfortunately, transparent objects often lack their  texture due to the reflection and refraction of light, which make them hard to be distinguished from the background.
The interactions between visible light and transparent objects are complex and sophisticated, and often hard to model. Generally, for many household transparent objects, the bulk of visible light passes directly through, and only a fraction ($4\%$ to $8\%$) \cite{DBLP:conf/cvpr/KalraTRVRK20}, depending on the refractive indices, is reflected. This often results in the background behind the  transparent object dominating the depth information of the objects.
\begin{figure}
	\centering
	\includegraphics[height=0.23\textheight]{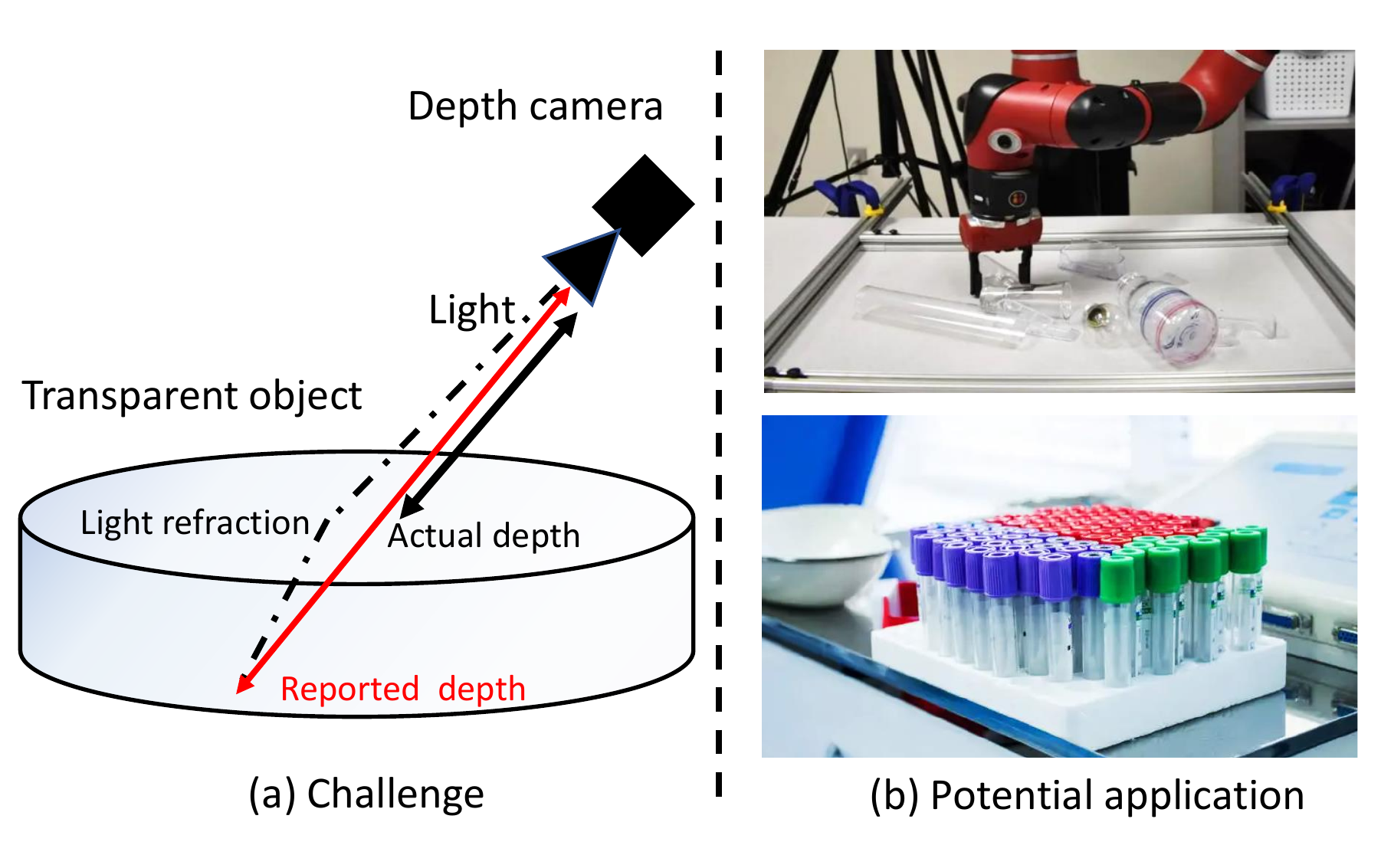}
	\caption{(a) shows the main challenges when estimating the depth of transparent objects. The transparency of the object's surface material causes a large bias in depth prediction by the depth camera. (b) shows some potential applications involving transparent objects such as object manipulation and instrument processing.  }
	\label{swin}
\end{figure}

Estimating the depth of transparent objects is critical for scene understanding, in which the depth estimation of transparent areas allows us  to achieve a  richer representation of the objects themselves and surroundings.         
However, the existing commercial-grade depth cameras, e.g.,  Intel RealSense \cite{keselman2017intel}, and Microsoft Kinect \cite{6547194}, still can not provide satisfactory depth perception of transparent objects. The reflection and refraction of light around transparent objects  lead to several challenges in the transparent object image processing tasks.  In scenes with dense clutter of transparent objects, the edges of objects are difficult to identify. In particular, in some extreme cases,  a few edges are even not visible at all.  

Consequently, there have been many attempts to  address the  depth estimation of transparent objects by deep neural networks. Eigen et al.\cite{eigen2014depth} are one of the first to use a multi-scale convolutional neural network (CNN) to directly predict the depth image from the color input. A  ResNet-based fully convolutional network architecture \cite{laina2016deeper} is introduced to regress the depth information. ClearGrasp\cite{9197518} provides a large-scale dataset of paired RGB-D transparent objects.  At present, a pretty intuitive approach is to train a fully-convolutional neural  network to directly regress the depth information.

Besides, the CNN-based transparent object depth estimation has a few limitations that the convolutional operation is local and the restricted receptive field leads to poorly extracted contextual information. Each convolutional kernel   concentrates more  on a sub-region of the entire image.  To enlarge the receptive field, CNN-based models  always keep stacking convolutional layers,  resulting in the  loss of global contextual information.
 Generally, transparent objects have no color and their appearance is often heavily influenced by background.  Therefore, the model should have a
stronger capability to extract contextual information and retrieve cues from the corresponding RGB and depth images.  
In fact, directly predicting the absolute depth from monocular images is challenging even for humans. Therefore the model must  be strong enough in  terms of contextual feature extraction.

	

In this paper, 
we gradually gather the token-based representation from different transformer layers in the encoder and progressively assemble them into a  full-resolution depth prediction.
The decoder receives multi-scale representation  at different stages with global receptive fields.  The fused feature representation is eventually  restored to  a resolution of the same size as the inputs for pixel-wise prediction.
We conduct extensive experiments on multiple datasets.
The performance of our model is improved by a large margin compared to previous best performing convolutional counterparts. Qualitative results indicate that our model can produce more fine-grained and globally coherent predictions.

The contribution of this paper can be summarized as follows:
\begin{itemize}
	\item  To the best of our knowledge, this is one of the first attempts to introduce  a vision transformer to perform depth estimation of transparent objects.

	\item TODE-Trans contains a contextual fusion module to better  perceive global information for further depth prediction and shows promising generalization to novel objects and unseen scenes.
	\item  Experimental results demonstrate the effectiveness of our model  on a range of testing datasets. The studies show that our model achieves competitive results compared with existing solutions and also exhibits promising generalization across different datasets.
\end{itemize}


\section{Related Work}

\textbf{Transparent depth prediction and estimation.} As the depth sensing has received increasing attention over the last   decade due to its  growing significance in autonomous driving, 3D vision, and  visual grasping.  Earlier method\cite{7298980} used hand-crafted approaches to complete the missing  values on the surface of objects in the depth image. 
Several studies usually assume that some information is already known, such as background texture \cite{DBLP:conf/cvpr/QianGY16}
or the 3D model of transparent objects \cite{DBLP:conf/icra/LysenkovR13}.
Previous work studies geometry-related properties  and some patterns unique to transparent objects. Many researchers \cite{DBLP:journals/tog/YeungTBK11},\cite{DBLP:journals/tog/WuZQG018} are interested in studying the  geometry of transparent objects for reconstruction and composition.

Recent methods \cite{eigen2014depth} model the depth sensing as a machine learning prediction problem and use deep neural networks to perform depth estimation.
The benefit of using deep neural networks for depth completion is that the model can learn from data for depth predictions rather than copying and interpolating from the input.
Recently, the majority of current works is still concerned with inaccurate depth estimation due to LIDAR noise in outdoor conditions such as Cityscapes \cite{DBLP:conf/cvpr/CordtsORREBFRS16} and KITTI \cite{DBLP:journals/ijrr/GeigerLSU13} datasets. 
Even in indoor environments,  depth images captured by the RGB-D camera  usually lose more than half of the pixels on objects that are too far or too close to the camera \cite{DBLP:conf/cvpr/ZhangF18}.  In addition, inferring the depth of transparent objects in the scene not only suffers from structural noise of the camera itself, but also from further interference arising from refraction of light by transparent materials.
This problem is further exacerbated by refraction on objects with too bright and glossy surface.

Currently, depth completion can be broadly categorized into two classes according to the input type.  The first class  directly regresses the depth map from the color image.
A number of approaches  \cite{DBLP:journals/pami/RanftlLHSK22},\cite{laina2016deeper},\cite{DBLP:conf/nips/ChenFYD16},\cite{eigen2014depth}, \cite{DBLP:conf/cvpr/FuGWBT18}
adopt convolutional neural networks to make deep predictions from RGB images. An alternative line of approaches \cite{DBLP:conf/cvpr/QiuCZZLZP19}, \cite{DBLP:conf/iccv/ChenYLU19},\cite{DBLP:conf/icra/MaK18} combines the RGB and sparse raw depth images to perform depth  prediction. Compared to the color-only input methods, these approaches improve the performance but still suffer from the generated low-quality  results due to the limited depth information. Recent studies \cite{DBLP:conf/cvpr/ZhangF18},\cite{9197518} employ deep networks to estimate the  occlusion boundaries and  surface normals from colored images and to perform prediction of transparent regions based on these estimations. However, if the model's predictions, e.g.  occlusion
 boundaries  are inaccurate, it may lead to poor results, and predicting multiple components would make the optimization process very slow.

\textbf{Transformer for dense prediction.}
Transformer \cite{DBLP:conf/nips/VaswaniSPUJGKP17} is originally developed for sequence to sequence tasks, such as machine translation, and  question answering.  
Later, owing to its excellent power to model the connections between tokens in a sequence,  the attention-based models, especially transformer, have become the governing paradigm in natural language processing (NLP). 
ViT\cite{dosovitskiy2020image} has achieved a further breakthrough by applying the standard transformer from NLP to computer vision. 
A visual-oriented transformer, Swin Transformer \cite{liu2021swin}, is proposed that integrates both the lone-range pixel modeling of attention mechanism and the inductive bias of convolutional kernels for local information. SETR \cite{DBLP:conf/cvpr/ZhengLZZLWFFXT021} models semantic segmentation with transformer encoder into a sequence-to sequence prediction task.
Wang  et al.\cite{wang2022transformer} present the use of transformer for visual grasping on a physical robotic arm, and their results show that the global  features extracted by  the transformer is more favorable for dense pixel predictions. However, these mentioned works are dedicated to outdoor scenes or other dense prediction tasks, few of them    concentrate on depth prediction for transparent objects.

\section{Problem Formulation}
Given a pair of the RGB scene image $I \in R^{3 \times H \times W} $ and inaccurate depth image $D \in R^{H \times W} $, the goal is to obtain accurate  depth predictions from monocular camera streams. 
Our method entails generating a clear and complete depth image  $\hat D \in R^{H \times W} $ for transparent objects, where H and W are the height and width of the input image.  The entire model is designed to learn a  function ${f}$ that  maps the RGB scene image $I$ and inaccurate depth image $D$  to a refined depth prediction $\hat{D}$, defined in $\hat{D}=f({I},{D})$.

\begin{figure*}[]
	\centering
	\includegraphics[height=0.43\textheight]{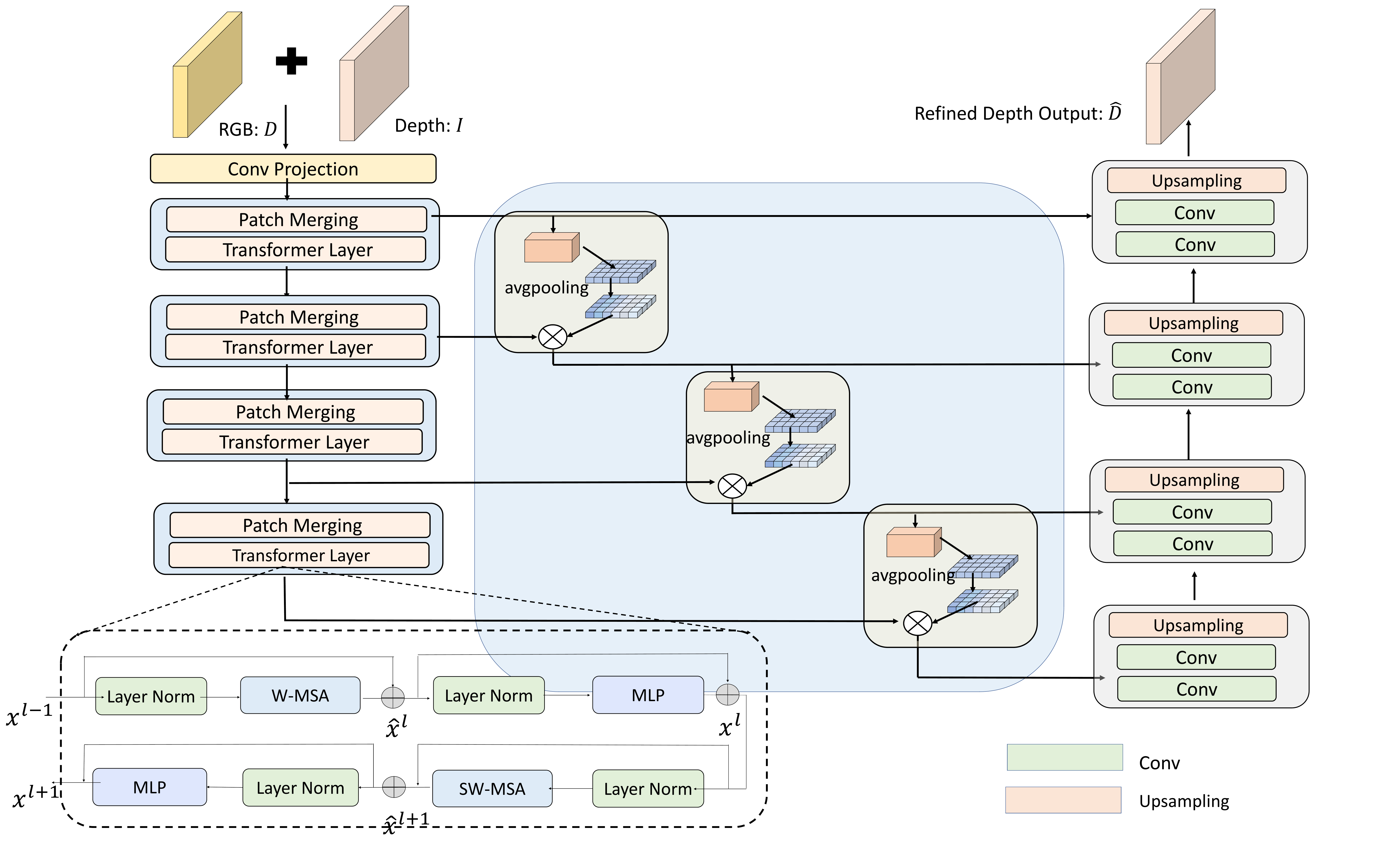}
	\caption{Overview of our proposed method. The model takes an RGB  $\mathcal{I} \in \mathbb{R}^{H\times W \times 3}$ and an inaccurate depth map $\mathcal{D}  \in \mathbb{R}^{H \times W}$ as input, and predicts the depth output in an end-to-end manner. The entire network is composed of three stages: the first is a global feature extraction encoder consisting of stacking transformer layers, the middle is a progressive depth refinement module, and the bottom is a decoder made of convolutional blocks that gradually receive previous multi-scale features to recover the depth information of transparent objects. }
	\label{arch}
\end{figure*}
\section{Method}
\textbf{Architecture}.
As shown in Fig. 2, TODE-Trans contains an \emph{encoder}, a \emph{ fusion feature module}, and a \emph{decoder}.
We adopt the transformer in the encoder stage to globally extract features. A fusion perception module is included to iteratively merge the captured features.
We leverage swin transformer block \cite{liu2021swin} as the backbone to achieve a nice trade-off between local fine-grained features and global contextual features in the encoder phase. Our encoder consists of four stages. The first stage is composed of a patch projection layer.
Concretely,  for an input RGB-D image, we first split it into non-overlapping square patches with patch size of $2$,  Then, the patches  are treated as tokens. And then they are flattened into vectors and fed into transformer blocks.  This kind of encoder allows us to develop multi-scale features and can be used by the following module. The attention mechanism globally learns the depth relationships from different parts of the input image. We desire  to conduct the depth completion using more global information rather than just local features from the  color or raw depth image. 
An overview of the swin transformer block is presented in the  bottom of Fig. \ref{arch}. In the transformer block, at each layer, a Layer Norm (LN) layer is first applied to the input feature $x$ before it enters the attention layer. The calculation flow is shown as follows:
\begin{equation}
\begin{aligned}
  \hat{\mathbf{x}}^{l} &=\operatorname{W-MSA}\left(\mathrm{LN}\left(\mathbf{x}^{l-1}\right)\right)+\mathbf{x}^{l-1}, \\
  \mathbf{x}^{l} &=\operatorname{MLP}\left(\mathrm{LN}\left(\hat{\mathbf{x}}^{l}\right)\right)+\hat{\mathbf{x}}^{l}, \\
  \hat{\mathbf{x}}^{l+1} &=\operatorname{SW-MSA}\left(\mathrm{LN}\left(\mathbf{x}^{l}\right)\right)+\mathbf{x}^{l}, \\
  \mathbf{x}^{l+1} &=\operatorname{MLP}\left(\mathrm{LN}\left(\hat{\mathbf{x}}^{l+1}\right)\right)+\hat{\mathbf{x}}^{l+1},
\end{aligned}
\end{equation}
where LN refers to the layer normalization, and MLP means a multi-layer perception  with two layers. The feature $x^{l-1}$ of previous layer is first passed through window attention (W-MSA) for local information attention calculation. 
Similarly, then features are put through sliding window multi-head attention (SW-MSA) to establish global feature correlations. The foundation of the swin transformer block is the multi-head attention. For the original swin-trainsformer, $X$ is the input to the multi-head attention. The input $X$ is mapped to the query, key, and value by linear transformation,
\begin{equation}
Q =X W_Q, K=X W_K, V=X W_V,
\end{equation}
where $W_Q,W_K,$ and $W_V$ refer to the learnable projection matrices.  Afterwards, the attention weights between features are calculated by the inner  product of query (K) and key (K),
\begin{equation}
     \text{Attention}({Q},{K},{V})= \operatorname{SoftMax}(\frac{{Q} {K}^T}{\sqrt{d}}+{B}) {V},
\end{equation}
where $B$ refers to the position encoding added to the embedding  in each attention layer and $\sqrt{d}$ is a scaling factor.

In Fig. \ref{arch},  we  show that such global multi-scale feature extraction contributes to the quality of the final depth prediction, which naturally yields global coherent and fine-grained  prediction.
Also, it becomes essential to further encode and fuse the geometric features of transparent objects and its surrounding spatial arrangements.
Specially, a depth feature fusion module is designed to enhance  the information fusion, taking the global geometric feature of transparent objects from the encoder for further refinement.  The entire FFM consists of three feature fusion sub-modules \cite{hu2018squeeze} where each  sub-module improves the fusion efficiency by modelling the correlation between feature channels.  The sub-module in FFM  performs a global average pooling on the feature map passed in from the previous layer in the encoder, scales the generated output to the range [0,1], and  multiplies this value as scale with the feature map of the next layer in the encoder. With this successive fusion operation, the fusion sub-module enhances the important features and attenuates the less important  by controlling the magnitude of the scale, thus making the extracted features more directional.

To reduce the number of parameters, a lightweight and efficient decoder is constructed to obtain the estimated depth map, which restores the features generated by 1$\times$ 1 convolution kernel to a size of $H \times W $. Similar to the encoder, the decoder consists of four stages. Each stage contains two convolution layers followed by one upsampling layer. Except for the last stage, the output of the lower stage and the corresponding output of our method together are used as input.  In fact, we found that the encoder incorporating the attention as well as the decoder with convolutional layers better can help the model  in  forecasting find-grained information about the surface of transparent objects. 

We formulate the transparent object depth estimation as a dense regression problem. The object function is defined as:
\begin{equation}
\mathcal{L}=\left\|\hat{\mathcal{D}}-\mathcal{D}^{*}\right\|^{2}+\beta ( 1-\cos \left\langle\hat{\mathcal{D}}_{h} \times \hat{\mathcal{D}}_{w}, \mathcal{D}_{h}^{*} \times \mathcal{D}_{w}^{*}\right\rangle ),
\end{equation}
where the first term is a reconstrction loss that minimizes the $L_2$ distance between the predicted depth $\hat{\mathcal{D}}$  and the ground truth $\mathcal{D}^{*}$. 
The second term is a regular term which penalizes the inconsistency in the predicted depth. In addition, $\beta$ is a hyper-parameter that is set to 0.01.  We train the model with a weighted loss to supervised regression of the depth value,
\begin{equation}
\mathcal{L}_{final}=\alpha \mathcal{L}_{masked}+\beta \mathcal{L}_{unmasked},
\label{final}
\end{equation}
where  $\mathcal{L}_{masked}$ denotes training  by masking the non-transparent regions using the labels provided by the dataset and $\mathcal{L}_{unmasked}$ indicates a pixel-wise training on the entire input image. The final object function $\mathcal{L}_{final}$ can guide the model to learn from color and raw depth images to produce accurate and clear results.


\section{Experiment}
\begin{figure}[]
	
	\center
	\subfigure{
		\includegraphics[width=0.35\textheight]{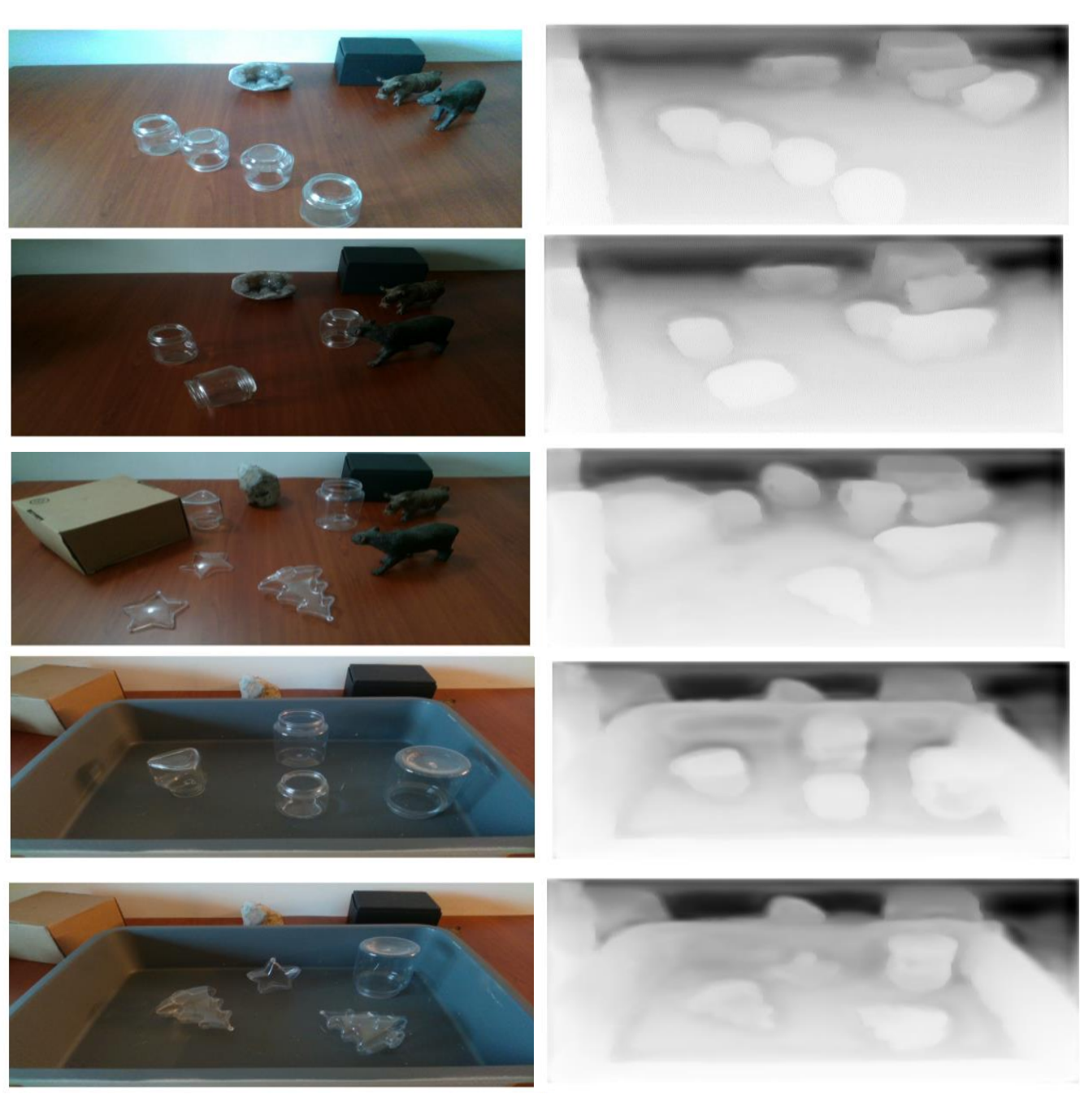}
	}

	\caption{  The samples of predicted depth images from our TODE-Trans model.  Under different lighting conditions and scenes,  our model provides effective depth prediction for  transparent objects.}	
	\label{sample}
\end{figure}
In this section,  we perform a series of experiments to evaluate our approach. Our investigation mainly concentrates on the following questions:   "How does the model retrieve clues for depth prediction from the colored image and inaccurate raw depth image?"; and "How well does the model generalize across datasets?".
\subsection{Datasets and evaluation metrics}
\textbf{Dataset.} The ClearGrasp \cite{9197518} and Omniverse Object dataset (OOD)  \cite{ZhuMXMEDF21} are used as the primary training and evaluation datasets. $\textbf{\romannumeral1}$) \emph{ClearGrasp}  training samples are generated by Synthesis AI’s platform which is composed of nine CAD models based on real-world transparent plastic objects. The total number of the training dataset is 23524.  For testing samples, ClearGrasp consists of synthesis data and real-world data. Each type of data all contains known data (i.e., the transparent objects exist in the training set) and novel data (i.e., the data  is not contained by training dataset).  
\begin{table}[]
	\centering
	\setlength\tabcolsep{1.0pt}
	{
		\small
		\caption{Comparison to previous transparent depth estimation methods where DA is an abbreviation for data augmentation and FFM is an abbreviation for feature fusion module. $\uparrow$ indicates the higher the better and  $\downarrow$ indicates the lower the better. }
		\begin{tabular}{c|cccccc}
			\hline
			Methods & RMSE${\downarrow}$ &REL${\downarrow}$& MAE $\downarrow$ & $\delta_{1.05}$ $\uparrow$ & $\delta_{1.10}$ $\uparrow$ & $\delta_{1.25}$ $\uparrow$  \\ \hline
			&\multicolumn{6}{c}{Cleargrasp Real-known} \\ \hline
			RGBD-FCN\cite{ZhuMXMEDF21} & 0.054 & 0.087& 0.048  & 36.32 & 67.11 & 96.26  \\ \hline
			NLSPN   \cite{ParkJHLK20} & 0.149 & 0.228& 0.127  & 14.04 & 26.67 & 54.32  \\ \hline
			
			CG      \cite{9197518} & 0.039 & 0.055& 0.029  & 72.62 & 86.96 & 95.58  \\ \hline
			
			LIDF-Refine \cite{ZhuMXMEDF21}& {0.028} & {0.033} & {0.020}  & {82.37} & {92.98} & {98.63}  \\ \hline
			TODE-Trans w/o DA  & 0.040 & 0.060& 0.033  & 58.21 & 79.91 & 98.15  \\ \hline
			TODE-Trans w/o FFM  & 0.024 & 0.031& 0.018  & 82.33 & 95.57 & 99.59  \\ \hline
			TODE-Trans (ours) & \textbf{0.021} & \textbf{0.026} & \textbf{0.015}  & \textbf{86.75} & \textbf{96.59} & \textbf{99.73}  \\ \hline
			
			&\multicolumn{6}{c}{Cleargrasp Real-novel} \\ \hline
			RGBF-FCN\cite{ZhuMXMEDF21} & 0.042 & 0.070& 0.037  & 42.45 & 75.68 & 99.02  \\ \hline
			NLSPN   \cite{ParkJHLK20} & 0.145 & 0.240 & 0.123  & 13.77 & 25.81 & 51.59  \\ \hline
			
			CG      \cite{9197518} & 0.034 & 0.045& 0.025  & 76.72 & 91.00 & 97.63  \\ \hline
			
			LIDF-Refine \cite{ZhuMXMEDF21}& {0.025} & {0.036} & {0.020}  & {76.21} & {94.01} & {99.35}  \\ \hline
			TODE-Trans w/o DA  & 0.025 & 0.041& 0.021  & 70.09 & 91.40 & 99.83  \\ \hline
			TODE-Trans w/o FFM  & 0.012 & 0.017& 0.009  & 95.54 & 98.94 & 99.96  \\ \hline
			TODE-Trans (ours) & \textbf{0.012} & \textbf{0.016} & \textbf{0.008}  & \textbf{95.74} & \textbf{99.08} & \textbf{99.96}  \\ \hline
		\end{tabular}
		\label{cleargrasp}
	}
\end{table}
To access the generalization of the model, we also evaluate the model on  TransCG\cite{9796631}.
$\textbf{\romannumeral2}$) \emph{TransCG} utilizes a robot to collect a new dataset based on real environments, which contains 57715 RGB-D images captured by two different cameras. The dataset contains 51 common objects in daily life that may lead to inaccurate depth images, which include reflective objects, transparent objects, translucent objects, and objects with dense small holes.

\textbf{Baselines.}
We use the \emph{RGBD-FCN} proposed in \cite{ZhuMXMEDF21} as the baseline. The method directly uses Resnet34  as the network structure to generate depth maps by direct regression. \emph{CG}\cite{9197518} denotes the ClearGrasp algorithm, which  adopts three networks to infer the surface normals, masked  transparent surfaces, occlusion, and contact edge around transparent surfaces. 
\emph{LIDF-Refine}\cite{ZhuMXMEDF21} is an abbreviation for  the local implicit depth function prediction followed by depth refinement and is the current best algorithm on ClearGrasp dataset. \emph{DFNet} \cite{9796631} is a lightweight variant of UNet. 

\textbf{Evaluation Metric.}
Following previous works\cite{9796631},\cite{ZhuMXMEDF21},\cite{DBLP:conf/cvpr/ZhangF18}, we choose 
Root Mean Squared Error (RMSE), Absolute Relative Difference (REL),  and Mean Absolute Error (MAE).

RMSE is a common indicator used to assess the quality  of predicted depth image, defined as follows:
$\sqrt{\frac{1}{|\hat{\mathcal{D}}|} \sum_{d \in \hat{\mathcal{D}}}\left\|d-d^{*}\right\|^{2}}$
where $d$ is the predicted depth, and $d^{*}$ is the ground truth.

REL is the average absolute  relative difference between  the predicted depth and the ground truth,
$\frac{1}{|\hat{\mathcal{D}}|} \sum_{d \in \hat{\mathcal{D}}}\left|d-d^{*}\right| / d^{*}$.

MAE is the mean absolute error:
$\frac{1}{|\hat{\mathcal{D}}|} \sum_{d \in \hat{\mathcal{D}}}\left|d-d^{*}\right|$.

Threshold $\delta$: the percentage of pixels for which the predicted depth satisfies 
$\max \left(\frac{d_{i}}{d_{i}^{*}}, \frac{d_{i}^{*}}{d_{i}}\right)<\delta$. In this paper, $\delta$ is set to 1.05, 1.10, and 1.25.
For indicators such as RMSE, REL, MAE, etc., the lower  the better, for threshold $\delta$ the higher the better.
\subsection{Implementation Details}
All experiments are conducted on an Intel I5-12333 CPU with 16 physical cores and an Nvidia RTX 3080Ti GPU. For the training process, we set the batch size to 16. And the resolution of images is  set to $320 \times 240$ during training and testing. In the process of training,  we use the AdamW optimizer to train our model with an initial learning rate of $1e-3$ and the weight decay of 0.05.
A multi-step learning rate schedule is utilized and the model   is trained for a total of 40 epochs. To prevent overfitting, a variety of data augmentation methods is used before training, such as random flipping, and rotating.
Concerning loss, we set $\beta = 0.01 $.
The embedding projection layer is implemented by convolutional operator with a kernel size of 2 with stride size 2.
The final upsampling layer is followed by a $1\times1$ convolution to align the output to the desired  depth image.

\begin{table}[]
	\centering
	\setlength\tabcolsep{2.0pt}
	{
		\small
		\caption{Comparison results on the TransCG dataset }
		\begin{tabular}{c|cccccc}
			\hline
			Methods & RMSE${\downarrow}$ &REL${\downarrow}$& MAE $\downarrow$ & $\delta_{1.05}$ $\uparrow$ & $\delta_{1.10}$ $\uparrow$ & $\delta_{1.25}$ $\uparrow$  \\ \hline

			CG \cite{9197518}        & 0.054 & 0.083& 0.037  & 50.48 & 68.68 & 95.28  \\ \hline
			LIDF-Refine\cite{ZhuMXMEDF21} & 0.019 &  0.034 & 0.015  & 78.22 & 94.26 & 99.80  \\ \hline
			DFNet\cite{9796631} & 0.018 & 0.027& 0.012  & 83.76 & 95.67 & 99.71  \\ \hline
			TODE-Trans (ours) & \textbf{0.013} & \textbf{0.019}& \textbf{0.008}  & \textbf{90.43} & \textbf{97.39 }& \textbf{99.81 } \\ \hline
		\end{tabular}
		\label{transcg}
	}
\end{table}
\begin{table}[]
	\centering
	\setlength\tabcolsep{1.5pt}
	{
		\small
		\caption{Comparison results on cross-domain  experiments. }
		\begin{tabular}{c|cccccc}
			\hline
			&\multicolumn{6}{c}{Train Clear+OOD Val TransCG} \\ 
			\hline
			Methods & RMSE${\downarrow}$ &REL${\downarrow}$& MAE $\downarrow$ & $\delta_{1.05}$ $\uparrow$ & $\delta_{1.10}$ $\uparrow$ & $\delta_{1.25}$ $\uparrow$  \\ \hline

			CG \cite{9197518}        & 0.061 & 0.108& 0.049  & 33.59 & 54.73 & 92.48  \\ \hline
			LIDF-Refine\cite{ZhuMXMEDF21} & 0.146 &  0.262 & 0.115  & 13.70 & 26.39 & 57.95  \\ \hline
			DFNet\cite{9796631} & 0.048 & 0.088& 0.039  & 38.65 & 64.42 & 95.28  \\ \hline
			TODE-Trans (ours) & \textbf{0.034} & \textbf{0.057}& \textbf{0.026}  & \textbf{64.10} & \textbf{78.86 }& \textbf{98.80 } \\ \hline
		\end{tabular}
		\label{cross-domain}
	}
\end{table}

\subsection{Comparison to State-of-the-art Methods}
Table  \ref{cleargrasp} shows our  results with the current state-of-the-art approaches on the ClearGrasp dataset. The empirical results reveal  that our proposed method is significantly better than other approaches on the real-known and real-novel transparent objects.  
For all compared baselines, we apply the same setting to guarantee a fair comparison.
For CG and LIDF-Refine, we use the released official codes and default hyper-parameters for training. For other methods, we present the results reported in their original papers.
As such, the methods that we consider for comparison include: (a) the classical full convolutional neural network scheme, (b) the current state-of-the art approaches in the transparent depth completion, (c) the 3D vision-based depth estimation of transparent objects, and (d) the convolution-based multi-scale feature fusion scheme.

For our approach, we use swin-transformer as the backbone network to recover the exact depth image. We show the point cloud  built from our predicted depth image, from which we can see that there is little difference between our results and ground truth. For example, in Fig. \ref{show_point_cloud}, the 3D point cloud of the transparent cups reconstructed from the original depth sensors misses most of the details below the top of cups. On the contrary, the whole transparent part is well reconstructed and completed by the prediction of our transformer model.
We also tested our algorithm on TransCG, and the results are shown in Table \ref{transcg}. Except for the results of our model, the results of the other algorithms are taken from the original paper \cite{9796631}. 
As shown in Tables  \ref{cleargrasp} and \ref{transcg}, our model achieves a significant improvement on all metrics compared to previous models.

\subsection{Results Discussion}
The low reflectivity of transparent materials can lead to inconsistent appearance in different backgrounds, which causes poor generalization of the model in novel environments.
We also conduct cross-domain experiments to show the robustness and generalization of our method. Table \ref{cross-domain} shows the cross-domain experimental results for training validation on different training and test sets. Similar to the experiments in TransCG, we perform the following  experiments:  training on ClearGrasp and OOD data  and then testing on TransCG.  The results show that our method has higher robustness and generalization than previous methods. In Table \ref{cross-domain},   most of these baselines are based on CNN and concentrate on local features, looking for patterns in the neighborhood pixels. Instead, the encoder in our model utilizes the attention mechanism to capture relationships from different parts in the input to recover the missing geometry.

\begin{figure}[]
	
	\center

	\includegraphics[width=0.37\textheight]{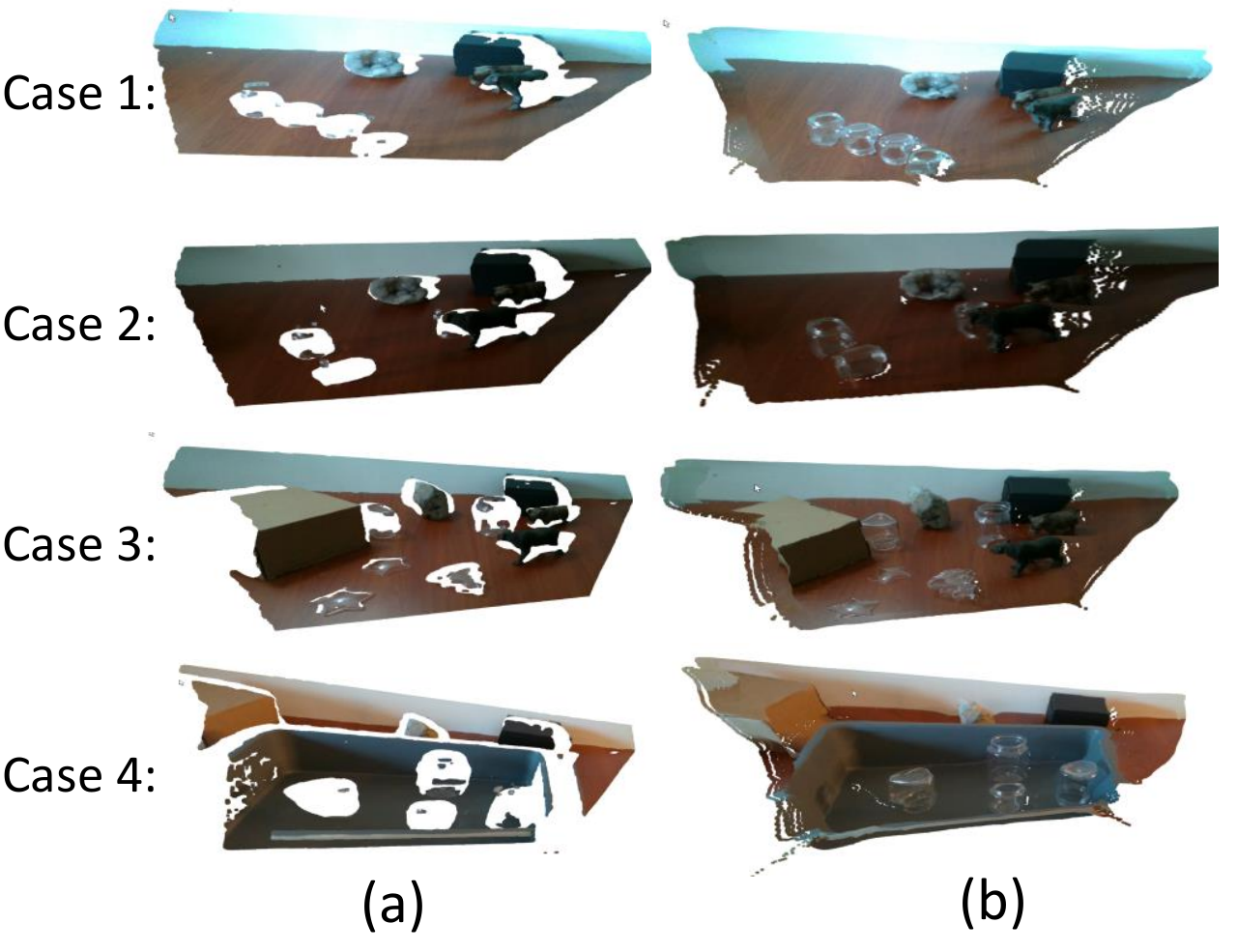}

	\caption{Overview of the 3D geometric appearance  of scene reconstructed by the depth and RGB images. (a)  shows the reconstructed 3D appearance of the original depth images. (b) shows the 3D point clouds reconstructed using our predicted depth images.  Compared to the original depth image, our predicted depth image achieves a better reconstruction and rendering result. For example, in both case1 and case2, TODE-Trans generates a clear point cloud for the glass in the figure under different lighting conditions.}	
	\label{show_point_cloud}
\end{figure}

First in the Table \ref{cleargrasp}, we compare the performance of different models in different situations, such as  real known data, real novel data, etc. For instance, when trained on real known data, the RMSE decreases from 0.028 to 0.021 with a $25\%$ improvement compared to LIDF-Refine.
Note that in all cases our model is better than the full convolutional neural network model (FCN). We conjecture that FCN suffers from information missing in the encoder stage by progressively downsampling during feature extraction so that the depth can not be recovered well in the decoder process.
Previous methods such as FCN do not perform  well, probably because the convolution operators ignore the dependencies between long-term features.

To help understand how the attention mechanism recovers the  geometric surface  of transparent objects, we also visualize the output of our model. In Fig. \ref{sample}, we can observe that our model maintains a clear boundary and refined detail.
The transparent object in the depth image often does not provide sufficiently accurate depth data, partially due to the  reflection of structured light by the surface of transparent objects. In Fig. \ref{show_point_cloud}, we can observe that more than one-third of the pixel information in the transparent object area of the raw depth image is missing. It is a great challenge for the model to recover  the depth information of the transparent area from the rest of the raw depth image and the paired RGB image, which demands a powerful ability to extract contextual information.

\begin{table}[]
	\centering
	\setlength\tabcolsep{4pt}
	{
		\small
		\setlength\tabcolsep{3.0pt}
		\caption{Quantitative comparison of different input modalities }
		\begin{tabular}{c|cccccc}
			\hline
			Methods & RMSE${\downarrow}$ &REL${\downarrow}$& MAE $\downarrow$ & $\delta_{1.05}$ $\uparrow$ & $\delta_{1.10}$ $\uparrow$ & $\delta_{1.25}$ $\uparrow$  \\ \hline

			Depth & 0.020 & 0.025 & 0.015  & 95.21 & 98.24 & 99.83  \\ \hline
			RGB & 0.30 & 0.39 & 0.23  & 9.52 & 18.69 &  40.96  \\ \hline
			Depth+RGB & 0.026 & 0.041& 0.021  & 74.89 & 89.95 & 98.59  \\ \hline
			
		\end{tabular}
		\label{rgb-d}
	}
\end{table}

\subsection{Ablation Studies}
In this section, to evaluate the quality of generated depth, we utilize the original depth with missing and inaccurate observations to  complete the depth, and the ground truth with RGB information of scene to render the surface reconstruction respectively.  In Fig. \ref{show_point_cloud}, we also visualize the 3D point clouds converted by RGB-D images for further comparison and each row displays one example. Each pixel in the depth image is rendered into a 3D point using camera intrinsics provided by the dataset, where the color information  is obtained from the corresponding RGB input.  The original depth image captured by the sensor is missing largely.  As a result, the 3D geometry of point clouds reconstructed using the raw depth has a lot of flaws and missing in many places. In contrast, the surface reconstruction by our predicted depth image contains less noise than the originals, and the geometric appearance of the point cloud is much clearer.

To investigate the effect of different modalities on the depth estimation of transparent objects, we train the model using RGB, depth, and RGB-D  as inputs, respectively in Table \ref{rgb-d}. In all of these, we adopt the same training procedure and hyper-parameter setting as described earlier.
Experimental results reveal that simply considering RGB or depth both ultimately degrades the final performance.
 Unlike the conventional depth estimation, for transparent objects,  simply using RGB or depth image alone would hurt the final predicted quality.
Also, we can find that our model presents a pretty sound balance in terms of the capacity and accuracy. After investigating which input modalities are useful for the transparent object depth completion, there are still many interesting questions to be addressed such as whether the fusion module is helpful for the  depth prediction and how the data augmentation contributes to the final improvements.
In Table \ref{cleargrasp}, we perform ablation experiments   without the fusion module  and the models are all trained from scratch. The qualitative and quantitative results demonstrate  that the fusion module contributes to the improved performance. Intuitively, fusing information from the global context of each stage is beneficial for the final depth prediction. From these results, it is evident that our proposed method can estimate the depth of transparent objects  well.

\section{Conclusions}
In this paper, we propose a model specially designed for depth completion of transparent objects, which  incorporates transformer and CNNs.  Our model also follows the encoder-decoder framework, in which the encoder leverages the transformer to extract fine-grained feature representation by modeling long-distance dependencies and global context information. The decoder gradually  aggregates features from various semantic scales via skip-connections, and then upsamples them to obtain the resulting depth estimation. The experimental results show that our model with a transformer makes it easier to extract contextual information for transparent object depth complementation, and shows good generalization on cross-domain datasets.


\ifCLASSOPTIONcaptionsoff
  \newpage
\fi

\end{document}